%% file: iSTS_IISCNLP.tex
\documentclass[11pt,letterpaper]{article}
\usepackage{naaclhlt2016}
\usepackage{times}
\usepackage{latexsym}
\usepackage{hyperref}

\usepackage{mathrsfs}
\usepackage{amssymb}
\usepackage{amsmath}
\usepackage{makecell}
\usepackage{graphicx}
\naaclfinalcopy % Uncomment this line for the final submission
\def\naaclpaperid{401} %  Enter the naacl Paper ID here

\title{IISCNLP at SemEval-2016 Task 2: Interpretable STS with ILP based Multiple Chunk Aligner}

 \author{Lavanya Sita Tekumalla \and Sharmistha \\
        lavanya.tekumalla@gmail.com sharmistha.jat@gmail.com \\ Indian Institute of Science, Bangalore, India}
\date{}

\begin{document}

\maketitle

\begin{abstract}
Interpretable semantic textual similarity (iSTS) task adds a crucial explanatory layer to pairwise sentence similarity. We address various components of this task: chunk level semantic alignment along with assignment of similarity type and score for aligned chunks with a novel system presented in this paper. We propose an algorithm, iMATCH, for the alignment of multiple non-contiguous chunks based on Integer Linear Programming (ILP). Similarity type and score assignment for pairs of chunks is done using a supervised multiclass classification technique based on Random Forrest Classifier. Results show that our algorithm iMATCH has low execution time and outperforms most other participating systems in terms of alignment score. Of the three datasets, we are top ranked for answer-students dataset in terms of overall score and have top alignment score for headlines dataset in the gold chunks track.

\end{abstract}

\section{Introduction and Related Work}
Semantic Textual Similarity (STS) refers to measuring the degree of equivalence in underlying semantics(meaning) of a pair of text snippets. It finds applications in information retrieval, question answering and other natural language processing tasks. Interpretable STS (iSTS) adds an explanatory layer, by measuring similarity across chunks of segmented text, leading to an improved interpretability. It involves aligning multiple chunks across sentences with similar meaning along with similarity score(0-5) and type assignment. 

%\section{Related Work}
Interpretable STS task was first introduced as a pilot task in 2015 Semeval STS task. Several approaches were proposed including NeRoSim \cite{sts:nerosim}, UBC-Cubes  \cite{sts:ubccubes} and  Exb-Thermis \cite{hanig-remus-delapuente:2015:SemEval}. For the task of alignment, these submissions used approaches based on monolingual aligner using word similarity and contextual features \cite{sultan2014}, JACANA that uses phrase based semi-markov CRFs \cite{jacana2013} and Hungarian Munkers algorithm \cite{hungarianmunkers}. Other popular approaches for mono-lingual alignment include two-stage logistic-regression based aligner \cite{sultan2015}, techniques based on edit rate computation such as \cite{houda2011} and TER-Plus \cite{terplus}. \cite{Bodrumlu:2009:NOF:1611638.1611642} used ILP for word alignment problem. The iSTS task in 2016 introduced problem of many-to-many chunk alignment, where multiple non-contiguous chunks of the source can align with multiple-non-contiguous chunks of the target sentence, that previous monolingual alignment techniques cannot handle. %Previous approaches in iSTS task were built to work with one-to-one alignment. 
We propose iMATCH, a new technique for monolingual alignment for many-to-many alignment at the chunk level, that can combine non-contiguous chunks based on integer linear programming (ILP). We also explore several features to define a similarity score between chunks to define the objective function for our optimization problem, similarity type and score classification modules.
To summarize our contributions: 
\begin{itemize}
 \item We propose a novel algorithm for monolingual alignment : iMATCH that handles many-to-many chunk level alignment, based on Integer Linear Programming. 
 \item We propose a system for Interpretable Semantic Textual Similarity: In the Gold-chunks track, our system is the top performer for the students-dataset and our alignment score is in that of the best two teams for all datasets.
\end{itemize}

\section{System for Interpretable STS}

Our system comprises of (a) alignment module, iMATCH  (section \ref{sec:milp-align}), (2) Type prediction module (section \ref{sec:type}) and (3) Score prediction module (section \ref{sec:score}). In the case of system chunks, there is an additional chunking module for segmenting input sentences into chunks. Figure 1 shows the block diagram of proposed system. 
%We first describe our system in the context of the Gold Chunks track. 
\begin{figure}[t]
\label{fig:dia}
\centering
\includegraphics[width=2.8in,height=2.4in]{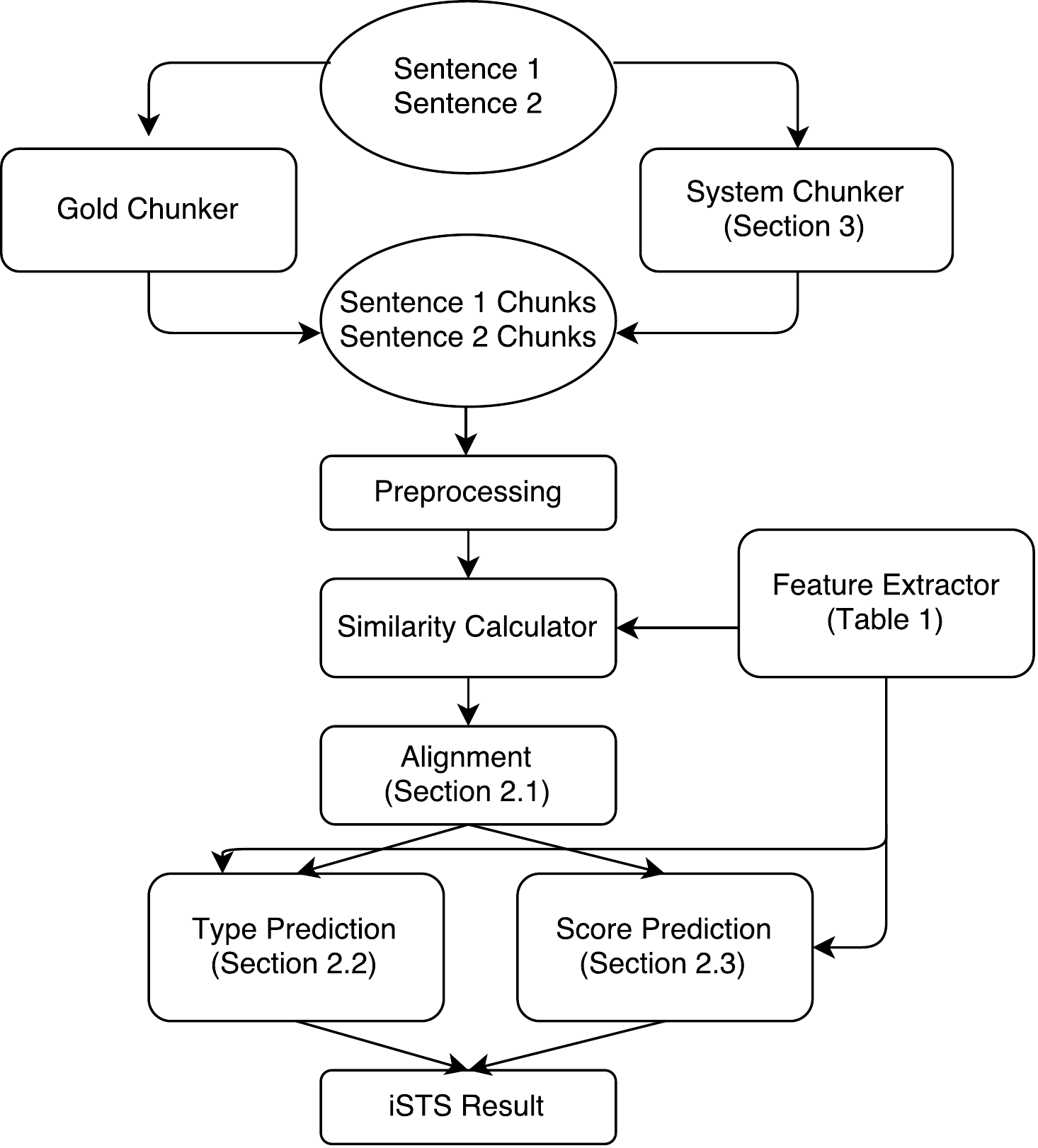}
\centering
\caption{Flow diagram of proposed iSTS system}
\vskip -1em
	\end{figure}
%---------- Alignement algorithm here
\input{ILP_algorithm}
%-----------
\subsection{Type Prediction Module}
	
{\scriptsize
\begin{table*}[t]
\label{tab:simi}
	\scriptsize
	\centering
	    \caption{Feature Extraction as used in various modules of iSTS system}
		\begin{tabular}{|l|l|l|}
		\hline
		{\bf No.} & {\bf Feature Name} & {\bf Description} \\\hline 
			\hline
			F1 & Common Word Count & \makecell{ $v1\_words =$ \{words1 from sentence 1\}  \\
		 $v2\_words =$ \{words2 from sentence 2\} \\
		feature value =  $\frac{\{|v1\_words \cap v2\_words|\} } {0.5*(sentence1\_length+sentence2\_length)}$ 	}
			\\
			
			\hline

			F2 & Wordnet Synonym Count & \makecell{ $v1 =$ \{words1\} $\cup$ \{wordnet synsets, similar\_tos, hypernms and hyponyms of words in sentence 1\} \\
		$v2 =$ \{words2\} $\cup$ \{wordnet synsets, similar\_tos, hypernms and hyponyms of words in sentence 2\}\\
		feature value =  $\frac{|v1 \cap v2|} {0.5*(sentence1\_length+sentence2\_length)}$ 	}
			\\
			
			\hline
			
			F3 & Wordnet Antonym Count & \makecell{ $v1\_words =$ \{words1\} ,$ v1\_anto =$  \{wordnet antonyms of words in sentence 1\} \\
		 $v2\_words =$ \{words2\} ,$ v2\_anto =$  \{wordnet antonyms of words in sentence 2\} \\
		feature value =  $\frac{\{|v1\_words \cap v2\_anto| \}+\{|v2-words \cap v1\_anto| \} } {0.5*(sentence1\_length+sentence2\_length)}$ 	}
			\\
			
			\hline
			
			F4 &\makecell{ Wordnet IsHyponym  \\ \& IsHypernym}& \makecell{ 
			$v1\_syn =$ \{synonyms of words in sentence 1\}, $v1\_hyp =$ \{hypo\{/hyper\}nyms of words in v1\_syn\} \\
			$v2\_syn =$ \{synonyms of words in sentence 2\}, $v2\_hyp =$ \{hypo\{/hyper\}nyms of words in v2\_syn\}	\\
			feature value = 1 if $|v1\_syn \cap v2\_hyp | \geq 1$			
			}		\\
			
			\hline
			F5 & Wordnet Path Similarity& \makecell{ 
			$v1\_syn =$ \{synonyms of words in sentence 1\}    \\
			$v2\_syn =$ \{synonyms of words in sentence 2\}	\\
			feature value = $\frac{\{ \sum{ w1.path\_similarity(w2)} \}}{(sentence1\_length+sentence2\_length)}, w1 \in v1\_syn, w2 \in v2\_syn  $	
			}		\\
			
			\hline
			F6 & Has Number& \makecell{ 
			feature value = 1 if phrase contains a number	
			}		\\
			
		%	\hline
		%	F7 & Adjective Count& \makecell{ 
		%	$v1\_adj =$ \{adjectives in sentence 1\}    \\
		%	$v2\_adj =$ \{adjectives in sentence 2\}	\\
		%	feature value =  $\frac{|v1\_adj| +|v2\_adj | } {max (sentence1\_length,sentence2\_length)}$
		%	}		\\
			
			\hline
			F7 & Is Negation& \makecell{ 
			feature value = 1 if one phrase contains a 'not' or 'n't' or 'never'  \\and other phrase does not contain those terms.	
			}		\\
			
			\hline
			
			F8 & Edit Score & \makecell{ 
			$v1 =$ words in sentence 1\\
			$v2 =$ words in sentence 2\\
			value = $\sum[ \max(1-\frac{EditDistance(w1,w2)}{(max(len(sentence1),len(sentence2))}), \forall w2 \in v2] \ \forall w1 \in v1 $. \\
			feature value = $\frac{value}{sentence1\_length}$\\
			For phrasal score, sum editscore of sentence 1 words with the closest sentence 2 words. \\Compute the average over scores for words in source.\\
			}		\\
			
			\hline
			F9 & PPDB Similarity & \makecell{ 
			$v1 = $words in sentence 1\\
			$v2 = $words in sentence 2 \\
			feature value = $\sum[ppdb\_similarity\{w1,w2\}], w1 \in v1, w2\in v2 $
			}		\\
			
			\hline
			F10 & W2V Similarity & \makecell{ 
			$v1 = $words in sentence 1, $v1\_vec =  \sum $word2vec embedding\{w1\}, w1 $\in$ v1  \\
			$v2 = $words in sentence 2, $v2\_vec =  \sum $word2vec embedding\{w2\}, w2 $\in$ v2  \\
			feature value = $cosine\_distance(v1\_vec,v2\_vec) $
			}		\\
			
			\hline
			F11 & Bigram Similarity & \makecell{ 
			$v1 = $bigrams in sentence 1,  \\
			$v2 = $bigrams in sentence 2, \\
			feature value = $cosine\_distance(v1,v2) $
			}		\\
			
			\hline
			F12 & Length Difference & \makecell{ 
			feature value = $len(sentence 1) - len(sentence 2) $
			}		\\
			
			\hline
		\end{tabular} 
		
	\end{table*}
}

\label{sec:type}
We use a supervised approach for multiclass classification based on the training data of 2016 and that of previous years (for some submitted runs) to learn the similarity type between aligned pair of chunks based on various features derived from the chunk text. We train a one-vs-rest random forest classifier \cite{scikit-learn} with various features mentioned in Table 1. We perform normalization on the input phrases as a preprocessing step before extracting features for classification. Normalisation step includes various heuristic steps to convert similar words to the same form, for example `USA' and `US' were mapped to `U.S.A'. Empirical results suggested that features F1, F2, F3, F5, F7, F8, F9, F12 along with unigram and bigram features give good accuracy with decision tree classifier. Feature vector normalisation is done before training and prediction. We note that our type classification module performed well for the answer-students dataset, while it did not generalize as well for the headlines and images. We are exploring other features to improve performance on these datasets as future work.
\subsection{Score Prediction Module}
\label{sec:score}
Similar to type classifier, we designed the Score classifier to do multiclass classification using one-vs-rest random forest classifier \cite{scikit-learn}. Each score 1-5 is considered as a class. `0' score is assigned by default for `not-aligned' chunks. Word normalization (US, USA, U.S.A are mapped to U.S.A string) is performed as a preprocessing step. Features  F1, F2, F3, F5, F7, F8, F9, F12 along with unigram and bigram features (refer Table 1) were used in training the multi-class classifier. Feature normalization was performed to improve results. Our score classifier works well on all datasets. The system achieved  highest score on the gold-chunks track for answer-students dataset and headlines dataset and is within 2\% of the top score for all other datasets.
%-------Sys chunks section
\input{sys_chunks_section}

%-------
\section{Experimental Results}
\label{sec:experiments}

\begin{table}[htb!]
		\scriptsize
		\centering
		\caption{Gold Chunks Images}
		\begin{tabular}{|l|l|l|l|l|l|}
			\hline
			{\bf RunName} & {\bf Align} & {\bf Type} & {\bf Score} & {\bf T+S} & {\bf Rank}\\\hline 
			IIScNLP R2& 0.8929 &0.5247 &0.8231 &0.5088& 15 \\
			IIScNLP R3 &0.8929 & 0.505  & 0.8264 & 0.4915 & 17 \\
			IIScNLP R1 & 0.8929 & 0.5015 & 0.8285 & 0.4845 & 19 \\  \hline
			UWB R3 & 0.8922 & 0.6867  & {\bf 0.8408} & 0.6708 & 1 \\ 
			UWB R1 & {\bf 0.8937} & 0.6829  & 0.8397 & 0.6672 & 2 \\ \hline
		\end{tabular} \\
		\caption{Gold Chunks headlines}
		\begin{tabular}{|l|l|l|l|l|l|}
			\hline
			{\bf RunName} & {\bf Align} & {\bf Type} & {\bf Score} & {\bf T+S} & {\bf Rank}\\\hline 
			IIScNLP R2& 0.9134 & 0.5755 & {\bf 0.829} & 0.5555 & 16 \\			
			IIScNLP R1 & {\bf 0.9144} & 0.5734 & 0.82 & 0.5509 & 18 \\ 
			IIScNLP R3 &0.9144 & 0.567 & 0.8206 & 0.5405 & 19 \\\hline
			Inspire R1 & 0.8194 & 0.7031 & 0.7865 & 0.696 & 1 \\ \hline
		\end{tabular} \\
		\caption{Gold Chunks Answer Students}
		\begin{tabular}{|l|l|l|l|l|l|}
			\hline
			{\bf RunName} & {\bf Align} & {\bf Type} & {\bf Score} & {\bf T+S} & {\bf Rank}\\\hline 
			IIScNLP R1& 0.8684 & {\bf 0.6511} & 0.8245 & {\bf 0.6385} & {\bf 1} \\			
			IIScNLP R2 & 0.8684 & 0.627 & {\bf 0.8263} & 0.6167 & 4 \\ 
			IIScNLP R3 & 0.8684 & 0.6511 & 0.8245 & 0.6385 & 2 \\  \hline
			V-Rep R2 & {\bf 0.8785} & 0.5823 & 0.7916 & 0.5799 & 8 \\ \hline
		\end{tabular} \\
		\caption{Gold Chunks Overall}
		\begin{tabular}{|l|l|l|l|l|l|}
			\hline
			{\bf RunName} & {\bf Images} & {\bf Headline} & {\bf Answer} & {\bf Mean} & {\bf Rank}\\
			&              & & {\bf Student }&& \\ \hline 			
			IISCNLP r2 & 0.5088 & 0.5555 & 0.6167 & 0.560 & 13 \\			
			IIScNLP R1 & 0.4845 & 0.5509 & {\bf 0.6385} & 0.558 & 14 \\ 
			IIScNLP R3 & 0.4915 & 0.5405 & 0.6385 & 0.557 & 15 \\ \hline
			UWB R1 & 0.6672 & 0.6212 & 0.6248 & 0.637 & 1 \\ \hline
		\end{tabular} \\
		%---------------------------------
		\caption{System Chunks Images}
		\begin{tabular}{|l|l|l|l|l|l|}
			\hline
			{\bf RunName} & {\bf Align} & {\bf Type} & {\bf Score} & {\bf T+S} & {\bf Rank}\\\hline  
			IIScNLP R2& {\bf 0.8459}& 0.4993 &0.777 &0.4872 &9 \\
			IIScNLP R3& 0.8335& 0.4862 &0.7654 &0.4744& 11 \\
			IIScNLP R1& 0.8335& 0.4862 &0.7654 &0.4744 &10 \\  \hline
			DTSim R3 & 0.8429 & 0.6276 & 0.7813  & 0.6095 & 1 \\ 
			Fbk-Hlt-Nlp R1 & 0.8427 & 0.5655 & {\bf 0.7862} & 0.5475 & 5 \\ \hline
			
		\end{tabular}\\
		
		\caption{System Chunks Headlines}
		\begin{tabular}{|l|l|l|l|l|l|}
			\hline
			{\bf RunName} & {\bf Alignment} & {\bf Type} & {\bf Score} & {\bf T+S} & {\bf Rank}\\\hline 
			IIScNLP R2& 0.821 & 0.508 & 0.7401 & 0.4919 & 9  \\
			IIScNLP R1& 0.8105 & 0.4888 & 0.723 & 0.4686 & 10  \\
			IIScNLP R3& 0.8105 & 0.4944 & 0.721 & 0.4685 & 11  \\  \hline
			DTSim R2 & 0.8366 & 0.5605 & {\bf 0.7595} & 0.5467 & 1 \\ 	
			DTSim R3 & {\bf 0.8376} & 0.5595 & 0.7586 & 0.5446 & 2 \\  \hline
		\end{tabular}\\
		\caption{System Chunks Answer Students}
		\begin{tabular}{|l|l|l|l|l|l|}
			\hline
			{\bf RunName} & {\bf Align} & {\bf Type} & {\bf Score} & {\bf T+S} & {\bf Rank}\\\hline 
			IIScNLP R3& 0.7563 & 0.5604 & 0.71 & 0.5451 & 2 \\
			IIScNLP R1& 0.756 & 0.5525 & 0.71 & 0.5397 & 5  \\
			IIScNLP R2& 0.7449 & 0.5317 & 0.6995 & 0.5198 & 6   \\  \hline
			Fbk-Hlt-Nlp R3 & {\bf 0.8166} & 0.5613 & 0.7574 & 0.5547 & 1 \\ 
			Fbk-Hlt-Nlp R1 & 0.8162 & 0.5479 & {\bf 0.7589} & 0.542 & 3  \\ \hline
		\end{tabular} \\
		
		\caption{System Chunks Overall}
		\begin{tabular}{|l|l|l|l|l|l|}
			\hline
			{\bf RunName} & {\bf Image} & {\bf Headline} & {\bf Answer} & {\bf Mean} & {\bf Rank}\\
			&              &  & {\bf Student }&& \\ \hline 	
			IISCNLP-R2 & 0.4872 & 0.4919 & 0.5198 & 0.499 & 8  \\
			IISCNLP-R3 & 0.4744 & 0.4685 & {\bf 0.5451} & 0.496 & 9  \\
			IISCNLP-R1 & 0.4744 & 0.4686 & 0.5397 & 0.494 & 10  \\ \hline
			DTSim R3 & 0.6095 & 0.5446 & 0.5029 & 0.552 & 1  \\ 	 \hline
		\end{tabular}
\end{table}

In this section, we present our results, in both the gold standard and the system chunks tracks. We submitted 3 runs for each track. 
In \emph{gold chunks track}, all three runs used the same algorithm, with different training data for the supervised prediction of type and score. While, in system chunks track, we submitted different algorithm and data combination for various runs. Detailed run information follows: \\
--\emph{ Gold Chunks - Run 1} uses training data from 2015+2016 for the headlines and images dataset and 2016 data for answer-students dataset.\\
--\emph{ Gold Chunks - Run 2} uses training data of all datasets combined from 2015 and 2016 for headlines and images and 2016 data for answer-students. \\
--\emph{ Gold Chunks - Run 3} uses 2016 training data alone for each dataset. \\
--\emph{System Chunks - Run 1} uses OpenNLP chunker with supervised training of type and score with data of all datasets combined for years 2015,2016.\\
--\emph{ System Chunks - Run 2} we use chunker based on stanford nlp parser and chunklink with training data of all datasets combined for years 2015,2016.\\
--\emph{ System Chunks - Run 3}, we use the OpenNLP chunker, with training data of 2016 alone.

Results of our system compared to the best performing systems in each track are listed in Tables 2-9. In both gold and system chunks track, run2 performs best owing to more data during training. Our system performed well for the answer-students dataset owing to our edit-distance feature that enables handling noisy data without any pre-processing for spelling correction. Our alignment score is best or close to the best in the gold chunks track, thus validating that our novel and simple approach based on ILP can be used for high quality monolingual multiple alignment at the chunk level. Our system took only 5.2 minutes for a single threaded execution on a Xeon 2420, 6 core system for the headlines dataset. Therefore, our technique is fast to execute. We observe that the quality of chunking has a huge impact on alignment and thereby the final score. We are actively exploring other chunking strategies that could improve results. Code for our alignment module is available at  {\url{https://github.com/lavanyats/iMATCH.git}}

\section{Conclusion and Future}
\label{sec:conclusion}
We have proposed a system for Interpretable Semantic Textual Similarity (task 2- Semeval 2016) \cite{task2-iSTS}. We introduce a novel monolingual alignment algorithm iMATCH for multiple-alignment at the chunk level based on Integer Linear Programming(ILP) that leads to the best alignment score in several cases. Our system uses novel features to capture dataset properties. For example, we designed edit distance based feature for answer-students dataset which had considerable number of spelling mistakes. This feature helped our system perform well on the noisy data of test set without any preprocessing in the form of spelling-correction. 

As future work, we are actively exploring features to improve our classification accuracy for type classification, which could help us improve out mean score. Some exploration in the techniques for simultaneous alignment and chunking could significantly boost the performance in sys-chunk track.  

\textbf{Acknowledgment} We thank Prof. Partha Pratim Talukdar, IISc for guidance during this research.  

\bibliography{iSTS16}
\bibliographystyle{naaclhlt2016}

\end{document}

%% file: ILP_algorithm.tex
{\bf Problem Formulation}: Following is the formal definition of our problem.
Consider source sentence ($Sent_1$) with M chunks and target sentence ($Sent_2$) with N chunks. 
Consider sets $C^1=\{ c_1^1,\hdots,c_M^1 \}$, the chunks of sentence $Sent_1$ and $C^2=\{ c_1^2,\hdots,c_N^2 \}$, the chunks of sentence $Sent_2$. Consider sets $\mathscr S_1 \subset PowerSet(C^1)-\phi$ and $\mathscr S_2 \subset PowerSet(C^2)-\phi$. Note that $\mathscr S_1$ and $\mathscr S_2$ are subsets of the power set (set of all possible combinations of sentence chunks) of $C^1$ and $C^2$ respectively. Consider sets $S_1 \in \mathscr S_1$ and $S_2 \in \mathscr S_2$, which denotes a specific subset of chunks that are likely to be combined during alignment. Let $concat(S_1 )$ denote the phrase resulting from concatenation of chunks in $S_1$ and $concat(S_2)$ denote the phrase resulting from concatenation of chunks of $S_2$. 
Consider a binary variable $Z_{S_1,S_2}$ that takes value 1 if $concat(S_1)$ is aligned with $concat(S_2)$ and 0 otherwise.

The goal of \emph{alignment module}  is to determine the decision variables ($Z_{S_1,S_2}$), which are non-zero. $S_1$ and $S_2$ can have more than one chunk (multiple alignment), that are not necessarily contiguous. Aligned chunks are further classified using Type classifier and Score classifier. \emph{Type prediction module} identifies a pair of aligned chunks ($concat(S_1), concat(s_2))$) with a relation type like EQUI (equivalent), OPPO (opposite) etc. \emph{Score classifier module} assigns a similarity score ranging between 0-5 for a pair of chunks. For the system chunks track, the \emph{chunking module}, converts sentences $Sent_1, Sent_2$  to sentence chunks $C_1,C_2$.

\subsection{iMATCH: ILP based Monolingual Aligner for Multiple-Alignment at the Chunk Level}
\label{sec:milp-align}
\begin{figure}
\centering
\fbox{	\includegraphics[width=7.9cm]{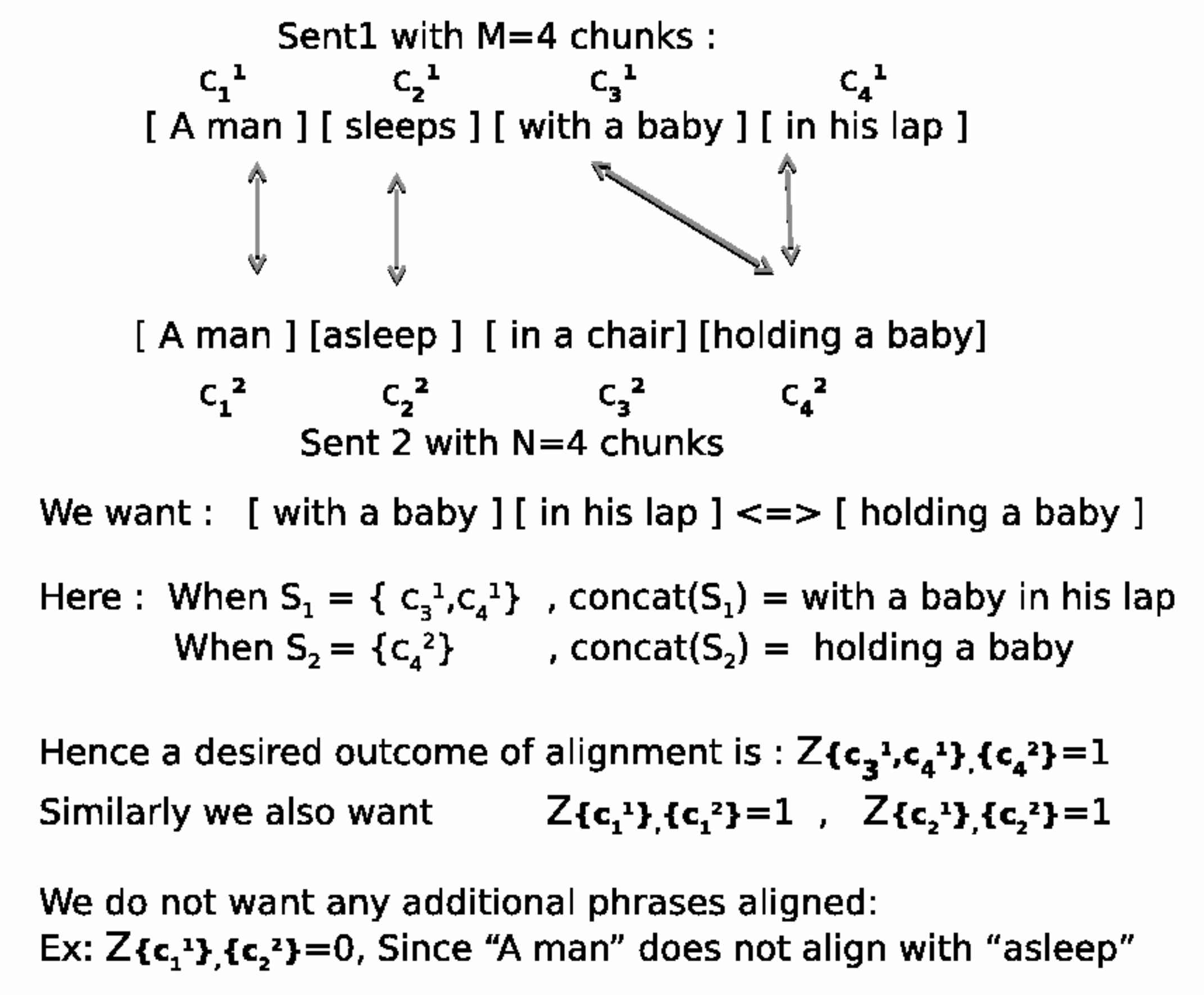}}
\caption{iMatch: An example illustrating notation}
\end{figure}
We approach the problem of multiple alignment (permitting non-contiguous chunk combinations) by formulating it as an Integer Linear Programming (ILP) optimization problem.
%We introduce variables $Z_{ij,kl}=1$ if (chunk i,chunk j) of source is aligned with (chunk k,chunk l) of target. 
We construct the objective function as the sum of all $Z_{S_1,S_2}, \forall S_1,S_2$ weighed by the similarity between $concat(S_1)$ and $concat(S_2)$, subject to constraints to ensure that each chunk is aligned only a single time with any other chunk. This leads to the following optimization problem based on Integer linear programming \cite{ilp}: 

{\begin{footnotesize}

\begin{equation}
\begin{aligned}
& \underset{Z}{\text{max}} 
 &  \underset{S_1 \in \mathcal S_1, S_2 \in \mathcal S_2} \Sigma Z_{S_1, S_2} ~ \alpha(S_1, S_2) ~ Sim({S_1, S_2}) \\ \nonumber
 & \text{S.T}\nonumber
 & \underset{\bar S_1 = \{S: c^1 \in S ,S \in \mathscr S_1 \}, S_2 \in \mathscr S_2 }{\Sigma} {Z_{S_1,S_2}} \leq 1, \forall 1 \leq c^1 \leq M\\ \nonumber
 & & \underset{S_1 \in \mathscr S_1, \bar S_2 = \{S: c^2 \in S, S \in \mathscr S_2 \} }{\Sigma} {Z_{S_1,S_2}} \leq 1, \forall 1 \leq c^2 \leq N \\ \nonumber
& &  Z_{S_1,S_2} \in \{0,1\}, \forall  S_1 \in \mathscr S_1,S_2 \in \mathscr S_2 \\\nonumber
\end{aligned}
\end{equation}
\end{footnotesize}}

\normalsize
Optimization constraints ensure that a particular chunk $c$ appears in an alignment a single time with any subset of chunks in the other sentence. Therefore, one chunk can be part of alignment only once. We note that all possible multiple alignments are explored by this optimization problem when 
 $\mathscr S_1 = PowerSet(C^1) - \phi$ and $\mathscr S_2 = PowerSet(C^2) - \phi$. However, this leads to a very high number of decision variables $Z_{S_1,S_2}$, not suitable for realistic use. Hence we consider a restricted usecase 
 
\normalsize
\begin{footnotesize}
$$\mathscr S_1 = \{C^1_1\}, \hdots, \{C^1_M\} \cup \{ \{C^1_i,C^1_j\}: 1\leq i < j \leq M\}$$ 
$$\mathscr S_2 = \{C^2_1\}, \hdots, \{C^2_N\} \cup \{ \{C^2_i,C^2_j\}: 1\leq i < j \leq N\}$$
\end{footnotesize}
\normalsize

This leads to many-to-many alignment where at most two chunks are combined to align with two other chunks. For iSTS task submission, we restrict our experiments to this setting (since this worked well for the iSTS task), but can relax sets $S_1$ and $S_2$ to cover combinations of 3 or more chunks. For efficiency, it should be possible to consider a subset of chunks based on adjacency information, existence of a dependency using dependency parsing techniques.
\normalsize
$Sim({S_1, S_2})$, the similarity score, that measures desirability of aligning $concat(S_1)$ with $concat(S_2)$, plays an important role in finding the optimal solution for the monolingual alignment task. We compute this similarity score by taking the maximum of similarity scores obtained from a subset of features F1, F2, F3, F8, F10 and F11 given in Table 1 as follows: $max(F1, F2, F3, F8, F10, F11)$. During implementation, the weighting term, $\alpha({S_1,S_2})$ is set as a function of the  cardinality of $S_1$ and cardinality of $S_2$ to ensure aligning fewer individual chunks (for instance, single alignment tends to increase objective function value more due to more aligned pairs, since similarity scores are normalized to lie between -1 and 1) does not get an undue advantage over multiple alignment. This is a hyper-parameter whose value is set using simple grid search. We solve the actual ILP optimization problem using PuLP \cite{pulp}, a python toolkit for linear programming. Our system achieved the best alignment score for headlines datasets in the gold chunks track.

% Our system achieved the best alignment score for headlines and images datasets in the gold chunks track. Student-answers dataset alignment score is also close to the best score submitted. Therefore, the proposed system for alignment is effective for the iSTS task. We note that the alignment results for the system-chunks track are highly dependent on the quality of chunking, for which we are exploring improved techniques.

%% file: sys_chunks_section.tex
\subsection{System Chunks Track: Chunking Module} 
\label{sec:syschunks}
When gold chunks are not given, we perform an additional chunking step. 
%of input sentences before subjecting them to alignment and type and score prediction. 
We use two methods for chunking: 
 (1) With OpenNLP Chunker\cite{opennlp}
 (2) With stanford-core-nlp \cite{stanfordnlp} API for generating parse trees and using the chunklink \cite{chunklink} tool for chunking based on the parse trees. 
 
 For chunking, we do preprocessing to remove punctuations unless the punctuation is space separated (therefore constitutes an independent word). We also convert unicode characters to ascii characters. Output of chunker is further post-processed to combine each single preposition phrase with the preceding phrase. We noted that the OpenNLP chunker ignored last word of a sentence, in which case, we concatenated the last word as a separate chunk. In the case of chunking based on stanford-core-nlp parser, we noted that in several instances, particularly in the student answer dataset, a conjunction such as `and' was consistently being separated into an independent chunk in most cases, and therefore improved chunking can be realized by potentially combining chunks around a conjunction. These processing heuristics are based on observations from gold chunks data. We observe that quality of chunking has a huge impact on the overall score in system chunks track. As future work, we are exploring ways to improve the chunking with custom algorithms.